\RequirePackage{etex}
\documentclass[10pt, twocolumn]{article}

\UseRawInputEncoding

\usepackage[pagenumbers]{cvpr} % To force page numbers, e.g. for an arXiv version
\usepackage{hyperref}  % Load this last

\usepackage{graphicx}
\usepackage{amsmath}
\usepackage{amssymb}
\usepackage{booktabs}
\usepackage{tikz}
\usepackage{svg}
\usepackage{wrapfig}
\RequirePackage{etex}
\usepackage{multirow}  % Required for multirow
\usepackage[capitalize]{cleveref}  % For easy cross-referencing

\usepackage[capitalize]{cleveref}
\crefname{section}{Sec.}{Secs.}
\Crefname{section}{Section}{Sections}
\Crefname{table}{Table}{Tables}
\crefname{table}{Tab.}{Tabs.}
\Crefname{figure}{Figure}{Figures}  % Capitalized for beginning of 
\crefname{figure}{Fig.}{Figs.}  % Lowercase for in-text use sentence

\begin{document}

\title{LSEAttention is All You Need for Time Series Forecasting}

\author{Dizhen Liang\\
{\tt\small liangdizhen@gmail.com}
}
\maketitle



\begin{abstract}
Transformer-based architectures have achieved remarkable success in natural language processing and computer vision. However, their performance in multivariate long-term forecasting often falls short compared to simpler linear baselines. Previous research has identified the traditional attention mechanism as a key factor limiting their effectiveness in this domain. To bridge this gap, we introduce LATST, a novel approach designed to mitigate entropy collapse and training instability—common challenges in Transformer-based time series forecasting. We rigorously evaluated LATST across multiple real-world multivariate time series datasets, demonstrating its ability to outperform existing state-of-the-art Transformer models. Notably, LATST, a single-layer Transformer, manages to achieve competitive performance with fewer parameters than some linear models on average, highlighting its efficiency and effectiveness.
\end{abstract}

\section{Introduction} \label{sec:intro}

Multivariate time series forecasting is crucial in diverse domains such as finance, healthcare, and environmental monitoring, where the goal is to predict future values based on historical data. This task is particularly challenging in long-term forecasting, as it necessitates models that can effectively capture feature correlations and long-term dependencies across multiple time series. Recent research has increasingly focused on applying Transformer architectures to time-series forecasting, leveraging their capacity to model complex temporal interactions through self-attention mechanisms. However, despite their potential, state-of-the-art approaches to multivariate time series forecasting still rely mainly on linear models, such as those proposed by Zeng\etal~\cite{Zeng2023}, raising concerns about the efficacy of Transformers in this context, particularly given their suboptimal performance and inherent complexity.

\subsection{Transformer-based Long-term Time Series Forecasting}

The application of Transformers to time series forecasting gained momentum following the introduction of the Transformer model  which established its potential as a universal backbone for various tasks across multiple modalities.\cite{Vaswani2017} The attention mechanism at the core of this architecture is formulated as follows:

\begin{equation} 
\text{Attention}(Q, K, V) = \text{Softmax}\left( \frac{Q K^T}{\sqrt{d_k}} \right) V, \label{eq:attention} 
\end{equation}

where \( Q \), \( K \), and \( V \) represent the query, key, and value matrices, respectively, and \( d_k \) denotes the dimension of the key vectors. This formulation allows Transformers to dynamically weigh the significance of various input elements, facilitating the capture of intricate dependencies within the data.

Recent studies have tailored Transformer models specifically for long-term time-series forecasting. For example, Informer introduces ProbSparse self-attention and distillation techniques to efficiently identify the most relevant keys to forecasting, thus reducing computational complexity and improving interpretability by focusing on essential temporal signals.\cite{Zhou2021} Similarly, Autoformer  employs decomposition and auto-correlation mechanisms, effectively merging classical and modern methodologies to enhance model performance.\cite{Wu2021}

FEDformer incorporates a Fourier-enhanced module that achieves linear complexity in both time and space, significantly improving scalability and efficiency for long sequences.\cite{Zhou2022} Instead of concentrating solely on point-wise attention, PatchTST emphasizes patch-level attention by treating patches, rather than individual time steps, as input units, allowing the model to capture richer semantic information across multiple time series - a critical aspect for effective long-term forecasting.\cite{Nie2023}

The challenges associated with applying Transformers to time series forecasting are further highlighted by Ilbert in their SAMformer model. SAMformer integrates Reversible Instance Normalization (RevIN) to address shifting data distributions and incorporates Sharpness-Aware Minimization (SAM) to improve training stability.\etal~\cite{Kim2022, Ilbert2024}This approach tackles critical issues, including entropy collapse and training instability, that have impeded Transformer models in this domain. 

\subsection{Entropy Collapse}

In computer vision and natural language processing, attention matrices can experience entropy or rank collapse, as demonstrated by Dong\etal~\cite{Dong2021}. This issue is exacerbated in time-series forecasting due to the frequent fluctuations inherent in temporal data, leading to substantial performance degradation. Ilbert \etal~\cite{Ilbert2024} identified the attention mechanism as a primary contributor to this problem and introduced SAM to mitigate it, thus enhancing model performance. However, the fundamental causes of entropy collapse remain inadequately addressed in the literature, warranting further exploration of its underlying mechanisms and effects on model performance.

\subsection{Contributions}

In this work, we present the following contributions:

\begin{itemize} 
    \item A novel theoretical framework linking entropy collapse with numerical stability, aimed at restoring the intended functionality of the attention for time-series data.
    \item Introduction of the LATST (LESAttention Time-Series Transformer) a Transformer with LSEAttention replacement that attempts to solve the entropy collapse issue, thereby enhancing the trainability and performance of Transformers for time series forecasting.
    \item Empirical validation of our approach on widely used multivariate long-term forecasting datasets, demonstrating that Transformers equipped with LSEAttention outperform state-of-the-art Transformers. 
\end{itemize}

%-------------------------------------------------------------------------
\section{Proposed Approach} 
\label{sec:pa}

\subsection{Problem Formulation}

In multivariate time series forecasting, the objective is to predict future values \( P \) for each channel, represented as \( Y \in \mathbb{R}^{C \times P} \). This prediction is based on historical time series data of length \( L \) across \( C \) channels, encapsulated in the input matrix \( X \in \mathbb{R}^{C \times L} \). The goal is to train a predictive model \( f_\omega: \mathbb{R}^{C \times L} \to \mathbb{R}^{C \times P} \), parameterized by \( \omega \), to minimize the mean squared error (MSE) between the predicted and actual values. The MSE is defined as:

\begin{equation}
\text{MSE} = \frac{1}{C} \sum_{c=1}^{C} \left\| Y_c - f_\omega(X_c) \right\|^2,
\end{equation}

where \( Y_c \) and \( X_c \) denote the true future values and historical input sequences for each channel \( c \).

\subsection{Motivation} \label{subsec:m}

Transformers have emerged as a powerful architecture for capturing temporal associations in sequential data, primarily due to their reliance on point-wise attention mechanisms. However, this reliance introduces vulnerabilities, including a phenomenon referred to as attention collapse, where attention matrices converge to nearly identical matrix. This identity diminishes the model's ability to differentiate patterns, thereby hindering generalization. An equally critical yet less-explored issue is entropy collapse, particularly prevalent in the application of Transformers to time-series forecasting. In such scenarios, Transformers might overfit to transient noise or abrupt changes in the data, resulting in significant performance degradation.\cite{Zeng2023} A deeper investigation into these phenomena reveals that the root cause often lies in the computation of the softmax function, especially the handling of its exponential terms.

In the attention mechanism \cref{eq:attention}, attention scores are computed as the dot product of query and key vectors. This operation tends to disproportionately emphasize abrupt changes or noise by assigning them larger values. Subsequently, when these scores are passed through the softmax function, defined as:

\begin{equation}
\text{softmax}(x_i) = \frac{e^{x_i}}{\sum_{j=1}^{n} e^{x_i}}, 
\label{eq:softmax}
\end{equation}

The exponential computation introduces significant numerical instability. Specifically, large attention scores lead to exponential overflow, while small attention scores result in underflow.\cite{Blanchard2019} This imbalance is further exacerbated during normalization, as the largest attention score (corresponding to the overflowed exponential value) disproportionately dominates the computation. Conversely, smaller attention scores, which result from underflowed exponential values, are effectively suppressed and contribute negligibly to the output. 

This disproportionate weighting undermines the diversity of the attention mechanism, reducing its capacity to effectively model complex temporal relationships. As a consequence, the model's ability to generalize is compromised, and its performance deteriorates in the presence of noisy or volatile data which is the inherent characteristic of time-series data. This behavior culminates in entropy collapse, where the attention distribution becomes overly concentrated on a few elements, thereby limiting the model's representational power. In other words, trend and periodicity might be barely considered.\cite{Zeng2023} Addressing this issue necessitates exploring alternatives to the conventional softmax function.
\begin{equation}
\text{LSE}(x) = \log\left(\sum_{i=1}^{n} e^{x_i}\right).
\label{eq:lse}
\end{equation}

To address these challenges, We propose a module termed LSEAttention, which integrates the Log-Sum-Exp (LSE) defined as \cref{eq:lse} trick proposed by Blanchard \etal~\cite{Blanchard2019} along with the Gaussian Error linear Unit (GELU) activation function introduced by Hendrycks \etal~\cite{Hendrycks2017}. The LSE trick alleviates entropy collapse issue arising from numerical instability through subtraction of max value. By starting from LSE where the softmax can be represented with LSE in this way:

\begin{equation}
\text{softmax}(x_i) = \frac{e^{x_i}}{\sum_{i=1}^{n} e^{x_i}} = e^{x_i - \text{LSE}(x)}.
\end{equation}

where \( e^{\text{LSE}(x)} \) denotes the exponential of the log-sum-exp function, enhancing numerical stability.

The LSE trick is stated as:

\begin{equation}
y = \log \left( \sum_{i=1}^{n} e^{x_i} \right) = \log \left( e^{a} \sum_{i=1}^{n} e^{x_i - a} \right)
\end{equation}

where \( a \) is a constant and max value utilized for ensuring that the largest exponent becomes zero and the others become non-positive. As a result, all exponential terms are bounded between 0 and 1 like normalization in practice. This can be simplified to:

\begin{equation}
y = a + \log \left( \sum_{i=1}^{n} e^{x_i - a} \right)
\label{eq:lse_simplification}
\end{equation}

To enhance the model's capabitlity, the GELU is applied to indroduce the non-linearity, then the equation is modfied as:

\begin{equation}
y = \text{GELU} \left( a + \log \left( \sum_{i=1}^{n} e^{x_i - a} \right) \right)
\label{eq:gelu_lse}
\end{equation}

This eventually can be expressed as the numerically stable manner of softmax function:

\begin{equation}
g_i = \exp(x_i - y) = \frac{e^{x_i}}{e^{y}} 
\label{eq:pre_softmax}
\end{equation}

\begin{equation}
\frac{e^{x_i}}{e^{\log \left( \sum_{i=1}^{n} e^{x_i} \right)}} = \frac{e^{x_i}}{\sum_{i=1}^{n} e^{x_i}} = \text{softmax}(x_i)
\label{eq:final_softmax}
\end{equation}

where \( g_i \) represents the \( i \)-th component of the stabilized softmax output.
Instead of strictly converting back to original softmax function in \cref{eq:final_softmax}.
We instead of applying softmax to the \( g_i \) to normalize the final attention weights into the bound between 0 and 1 in \cref{eq:norm_softmax}. 

\begin{equation}
\text{softmax}(g_i)
\label{eq:norm_softmax}
\end{equation}

In traditional Transformer architectures, the ReLU (Rectified linear Unit) activation function used in the Feed-Forward Network (FFN) is susceptible to the "dying ReLU" problem, where neurons can become inactive by outputting zero for all negative input values. This leads to a zero-gradient scenario for those neurons, effectively stalling their learning process and contributing to instability during training.

To address these challenges, Parametric ReLU (PReLU) is employed as an alternative activation function for Feed-Forward Network. PReLU introduces a learnable slope for negative inputs, allowing for a non-zero output even when the input is negative. This adaptation not only alleviates the dying ReLU problem but also facilitates a smoother transition between negative and positive activations, thereby enhancing the model's ability to learn from all regions of the input space \cite{He2015}. The presence of a non-zero gradient for negative values promotes improved gradient flow, which is crucial for training deeper architectures. Consequently, the use of PReLU contributes to overall training stability and helps maintain active representations, ultimately leading to enhanced model performance.

\subsection{LATST: Overall Structure}

\begin{figure}[ht]
\vskip 0.1in
\begin{center}
\centerline{\includegraphics[width=0.5\linewidth]{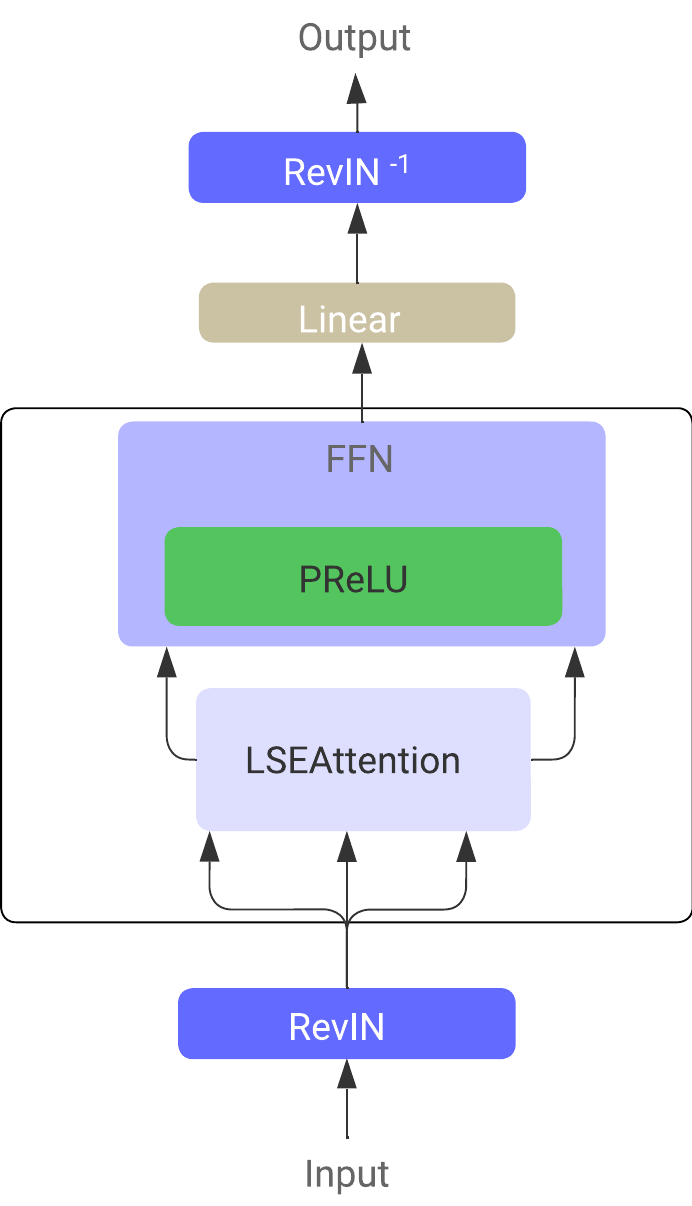}}
\caption{Overall structure of LATST.}
\label{fig:LATST}
\end{center}
\vskip -0.1in
\end{figure}

The proposed LSEAttention Time-Series Transformer (LATST) \cref{fig:LATST} builds upon the modifications discussed in Motivation \cref{sec:pa}, introducing a key enhancement: Reversible Instance Normalization. This normalization technique is particularly effective for mitigating discrepancies between training and testing data distributions in time series forecasting tasks as in \cite{Ilbert2024}

The architecture replaces the traditional temporal self-attention mechanism, with the LSEAttention module. To assess the efficacy of the LSEAttention component, we conducted an ablation study comparing LATST with the Gaussian Error linear Unit (GeLU) in the Feedforward Neural Network (FFN) against a variant utilizing the Rectified linear Unit (ReLU). The details of this study can be found in Ablation Study of \cref{sec:exp}.

Overall, the LATST architecture consists of a single-layer Transformer framework augmented with substitution modules, allowing adaptive learning while maintaining the robustness of attention mechanisms. This design facilitates effective modeling of temporal dependencies and enhances performance in time series forecasting tasks.

\subsection{LATST: Fractional Temporal Block Attention Encoder}
The feactional meaning is to individually build explicit temporal and variate transformer encoder.

The current approach of the time-series modelling by transformer relies on the the inplicit temporal modelling by mlp due to the impossible full division by traditional patching To tackle this problem by slicing 64 timesteps of input with all variables into a block, and the temporal encoder is used to model the explicit temporal relationship with absolute positional encoding.

The variate encoder is by default follow the paradim of iTransformer.

%------------------------------------------------------------------------
\section{Experiments}
\label{sec:exp}

\subsection{Dataset}
We conduct experiments on eight publicly available datasets of real-world multivariate time series that are widely recognized for long-term forecasting. These datasets include four Electricity Transformer Temperature datasets: ETTh1, ETTh2, ETTm1, and ETTm2, as well as the Electricity, Weather and Traffic datasets are sourced from Wu \etal~\cite{Wu2021}. The Exchange dataset is from Lai \etal~\cite{Lai2019}.  Each time series is segmented based on a specified input length to facilitate effective modeling.

\textbf{Sequence and Prediction Length:} In accordance with prior works, We set the input sequence length at 512 time steps for all models. The prediction length, which indicates the number of future time steps for which We aim to forecast data, varies among the set \{96, 192, 336, 720\}. This flexibility in prediction length allows for an evaluation of the model's performance across different forecasting horizons.

\textbf{Evaluation Metric:} Consistent with previous studies, we employ the widely used evaluation metric, mean squared error (MSE) and mean absolute error (MAE). This metric has been a standard choice in earlier works.

\subsection{Baseline}
We compare the performance of LATST against leading Transformer-based models, including SAMFormer \cite{Ilbert2024}, TSMixer \cite{Chen2023}, iTransformer \cite{Liu2024}, and PatchTST \cite{Nie2023}. 

\textbf{Hyperparameters:} For all experiments, the hyperparameter settings for LATST are derived from SAMFomer, with a batch size fixed at 8 for simplicity. Rest of the hyparameters could be referred in \cref{tab:class_dataset} where it demonstrates clear pattern between the set of parameters and the categories.

\begin{table}[t]
\centering
\caption{Classifications of Datasets with corresponding combination of hyperparameters where all are single-layer with Low Variates: \(N\) = 4, \(D\) = 32, \(F\) = 64; Mid Variates:\(N\) = 8 \(D\) = 32, \(F\) = 128; High Variates: \(N\) = 16, \(D\) = 128, \(F\) = 256; Ultra-High Variates: \(N\) = 16 \(D\) = 256, \(F\) = 512. It demonstrates a clear positive relationship between number of variates and number of heads, as well as the dimension sizes to better capture the channel interaction.}
\label{tab:class_dataset}
\small
\setlength{\tabcolsep}{8pt}  
\begin{tabular}{lc}
\toprule
\textbf{Dataset Type} & \textbf{Single-Layer, LR: 0.0001} \\
\midrule
\multirow{5}{*}{Low Variates} & \text{ETTH1} \\ & \text{ETTH2} \\ & \text{ETTM1} \\ & \text{ETTM2} \\ & Exchange \\ 
\multirow{1}{*}{\centering \text{Mid Variates}} & \text{Weather} \\
\multirow{1}{*}{\centering \text{High Variates}} & \text{ECL} \\
\multirow{1}{*}{\centering \text{Ultra-High Variates}} & \text{Traffic} \\
\bottomrule
\end{tabular}
\vskip -0.1in
\end{table}

\renewcommand{\arraystretch}{1.1} % Adjust row height if necessary
\begin{table*}[t]
\caption{Performance metric - MSE (mean squared error) for various time series forecasting models across multiple datasets and prediction horizons. The best results are highlighted in bold. LATST denotes the LSEAttention Time-Series Transformer model utilized in this study, while most results for other models are sourced from Ilbert \etal~\cite{Ilbert2024} with some ECL results from triplet layers Transformer on average from PatchTST \etal~\cite{Nie2023} and iTransformer \etal~\cite{Liu2024}.}  
\label{tab:performance}
\vskip 0.15in
\begin{center}
\begin{small}
\begin{sc}
\begin{tabular}{cccccccccc}
\toprule
\multicolumn{2}{c}{\textbf{Methods}} & \textbf{LATST}  & \textbf{SAMformer} & PatchTST &  \textbf{Transformer} & \textbf{TSMixer}   & iTransformer  & \textbf{Dlinear} \\
\midrule

\multirow{4}{*}{\parbox{0.5cm}{\centering \rotatebox{90}{\textbf{ECL}}}} 
% \multirow{4}{*}{\parbox{0.5cm}{\centering\textbf{Electricity}}} 
    & 96   & \textbf{0.129}& 0.155 & 0.129 & 0.182 & 0.173 & 0.148 & 0.140 \\ 
    & 192  & \textbf{0.150} & 0.168 & 0.147 & 0.202 & 0.204 & 0.162 & 0.153 \\ 
    & 336  & \textbf{0.168}& 0.183 & 0.163 & 0.212 & 0.217 & 0.178 & 0.169\\ 
    & 720  & 0.205 & 0.219 & 0.197 & 0.238 & 0.242 & 0.225 & \textbf{0.203}\\  \hline

\multirow{4}{*}{\parbox{0.5cm}{\centering \rotatebox{90}{\textbf{Traffic}}}} 

    & 96   & \textbf{0.369}& 0.407 & 0.462 & 0.420 & 0.409 & 0.395 & 0.410\\ 
    & 192  & \textbf{0.386}& 0.415 & 0.466  & 0.441 & 0.637 & 0.417 & 0.423\\ 
    & 336  & \textbf{0.397}& 0.421 & 0.482 & 0.501 & 0.747 & 0.433 & 0.436\\ 
    & 720  & \textbf{0.433}& 0.456 & 0.514 & 0.468 & 0.688 & 0.467 & 0.466\\ \hline

\multirow{4}{*}{\parbox{0.5cm}{\centering \rotatebox{90}{\textbf{Weather}}}} 
    & 96   & \textbf{0.146}& 0.197 & 0.177 & 0.227 & 0.214 & 0.174 & 0.176\\ 
    & 192  & \textbf{0.193}& 0.235 & 0.225 & 0.256 & 0.231 & 0.221 & 0.220\\ 
    & 336  & \textbf{0.246}& 0.276 & 0.278 & 0.278 & 0.279 & 0.278 & 0.265\\ 
    & 720  & \textbf{0.314}& 0.334 & 0.354 & 0.353 & 0.343 & 0.358 & 0.323\\ \hline

\multirow{4}{*}{\parbox{0.5cm}{\centering \rotatebox{90}{\textbf{ETTh1}}}} 

    & 96   & 0.381 & 0.381           & 0.414 & 0.509 & 0.398 & 0.386 & \textbf{0.375}\\ 
    & 192  & 0.415 & 0.409           & 0.460 & 0.535 & 0.426 & 0.441 & \textbf{0.405}\\ 
    & 336  & 0.434 & \textbf{0.423}  & 0.501 & 0.570 & 0.435 & 0.487 & 0.439\\ 
    & 720  & 0.471  & \textbf{0.427} & 0.500 & 0.601 & 0.498 & 0.503 & 0.472\\ \hline

\multirow{4}{*}{\parbox{0.5cm}{\centering \rotatebox{90}{\textbf{ETTh2}}}} 

    & 96   & \textbf{0.280} & 0.295          & 0.302 & 0.396 & 0.308 & 0.297 & 0.289\\ 
    & 192  & 0.346 & \textbf{0.340} & 0.388 & 0.413 & 0.352 & 0.380 & 0.383\\ 
    & 336  & 0.384 & \textbf{0.350} & 0.426 & 0.414 & 0.360 & 0.428 & 0.448\\ 
    & 720  & 0.423 & \textbf{0.391} & 0.431 & 0.424 & 0.409 & 0.427 & 0.605\\ \hline

\multirow{4}{*}{\parbox{0.5cm}{\centering \rotatebox{90}{\textbf{ETTm1}}}} 

    & 96   & \textbf{0.299} & 0.329 & 0.329 & 0.384 & 0.336 & 0.334 & \textbf{0.299}\\ 
    & 192  & 0.341 & 0.353           & 0.367 & 0.400 & 0.362 & 0.377 & \textbf{0.335}\\ 
    & 336  & \textbf{0.367} & 0.382           & 0.399 & 0.461 & 0.391 & 0.426 & 0.369\\ 
    & 720  & \textbf{0.420} & 0.429          & 0.454 & 0.463 & 0.450 & 0.491 & 0.425\\ \hline

\multirow{4}{*}{\parbox{0.5cm}{\centering \rotatebox{90}{\textbf{ETTm2}}}} 
    & 96   & 0.174  & 0.181          & 0.175 & 0.200 & 0.211 & 0.180 & \textbf{0.167}\\ 
    & 192  & 0.238 & 0.233          & 0.241 & 0.273 & 0.252 & 0.250  & \textbf{0.224}\\ 
    & 336  & 0.287 & 0.285 & 0.305 & 0.310 & 0.303 & 0.311 & \textbf{0.281}\\ 
    & 720  & \textbf{0.369}  & 0.375 & 0.402   & 0.426 & 0.390 & 0.412 & 0.397\\ \hline

\multirow{4}{*}{\parbox{0.5cm}{\centering \rotatebox{90}{\textbf{Exchange}}}} 
    & 96   & 0.102 & 0.161 & 0.088 & 0.292 & 0.343 & 0.086 & \textbf{0.081}\\ 
    & 192  & 0.201 & 0.246 & 0.176 & 0.372 & 0.342 & 0.177 & \textbf{0.157}\\ 
    & 336  & 0.395 & 0.368 & 0.301 & 0.494 & 0.484 & 0.331 & \textbf{0.305}\\ 
    & 720  & 0.935 & 1.003 & 0.901 & 1.323 & 1.204 & 0.847& \textbf{0.603}\\ 
    
\bottomrule
\end{tabular}
\end{sc}
\end{small}
\end{center}
\vskip -0.1in
\end{table*}

\subsection{Performance Analysis}
LATST demonstrates exceptional performance, achieving the lowest mean squared error (MSE) scores in 21 out of 32 scenarios across eight diverse datasets. This remarkable consistency underscores the model’s robustness and adaptability, effectively addressing a wide range of time-series forecasting tasks. Specifically, LATST outperforms all other Transformer-based models on the Electricity, Traffic, and Weather datasets across all evaluated metrics, demonstrating its superior ability to manage complex temporal dependencies.

When comparing LATST directly with SAMformer, the model shows its superiority by outperforming SAMformer on 6 out of 8 datasets with overall improvement by 4\% . Moreover, LATST surpasses the traditional Transformer model by a notable margin of 19.75\%, which emphasizes its enhanced predictive capabilities. Key innovations in LATST, such as the integration of the Log-Sum-Exp trick and refined attention mechanisms, contribute significantly to this performance enhancement. These modifications enable LATST to capture long-range dependencies and temporal patterns more effectively, setting it apart from its counterparts.

The detailed performance metrics for each dataset and prediction length are summarized in \cref{tab:performance}. Notably, LATST consistently achieves the lowest MSE across various time-series datasets that exhibit substantial variability, which highlights its capacity to effectively exchange information between different variables. This phenomenon is closely related to the number of variables and parameter settings, where the model’s performance improves proportionally with the increasing complexity of interactions. A more in-depth analysis can be found in \cref{tab:parms}.

In comparison with DLinear, LATST outperforms the model on 5 out of 8 datasets, despite DLinear having 139.7K parameters. LATST, with fewer parameters in most scenarios, demonstrates that its performance is not solely dependent on model size, but rather on the efficiency and effectiveness of its architecture.

\textbf{Consistency Across Datasets:} LATST’s ability to achieve the best MSE scores in a significant number of scenarios further reinforces its robustness and generalizability across diverse forecasting challenges. This performance stability across different datasets suggests that LATST effectively learns and captures the underlying temporal patterns, allowing it to adapt well to varying data distributions and dynamics. For instance, in the Traffic dataset, LATST consistently maintains lower MSE values even at longer prediction horizons, illustrating its resilience against overfitting and its capacity to generalize beyond training conditions.

\begin{table}[t]
\begin{center}
\centering
\caption{Performance of LATST with different activation functions for prediction length of 720.}
\vskip 0.15in
\begin{small}
\begin{sc}
\begin{tabular}{ccc}
\toprule
\textbf{Dataset} & \textbf{With PReLU} & \textbf{With ReLU} \\
\midrule
ECL & 0.205 & 0.205 \\
Traffic & \textbf{0.433} & 0.437 \\
Weather & 0.314 & 0.314 \\
ETTh1 & 0.471 & 0.471 \\
ETTh2 & \textbf{0.423} & 0.426 \\
ETTm1 & \textbf{0.420} & 0.423 \\
ETTm2 & 0.369 & 0.369 \\
Exchange Rate & \textbf{0.935} & 1.043 \\
\bottomrule
\label{tab:ab_relu}
\end{tabular}
\end{sc}
\end{small}
\end{center}
\vskip -0.1in
\end{table}

\subsection{Ablation Study Note Updated Yet}
\label{subsec:as}

In this section, we present a comprehensive ablation study aimed at quantitatively assessing the impact of various architectural components within the LATST framework. By systematically removing or modifying key elements, such as the GeLU and the PReLU, their contributions are isolated to the overall performance of the model. For consistency, we fix the experimental parameters to a prediction length of 96 and utilize four datasets: ETTh2, ETTm1, ETTm2, and Exchange Rate.

\begin{table}[t]
\begin{center}
\centering
\caption{Performance of LATST without GeLU compared to Transformer for prediction length of 192.}
\vskip 0.15in
\begin{small}
\begin{sc}
\begin{tabular}{ccc}
\toprule
\textbf{Dataset} & \textbf{With GeLU} & \textbf{Without GeLU} \\
\midrule

ECL & \textbf{0.150} & 0.153 \\
Traffic & \textbf{0.386} & 0.393 \\
Weather & \textbf{0.193} & 0.204 \\
ETTh1 & \textbf{0.415} & 0.433 \\
ETTh2 & \textbf{0.346} & 0.378 \\
ETTm1 & \textbf{0.341} & 0.357 \\
ETTm2 & \textbf{0.238} & 0.246 \\
Exchange Rate & \textbf{0.201} & 0.236 \\

\bottomrule
\label{tab:ab_gelu}
\end{tabular}
\end{sc}
\end{small}
\end{center}
\vskip -0.1in
\end{table}

\begin{table}[t]
\begin{center}
\centering
\caption{Overall performance improvement and average size of parameters with data of other models sourced from Ilbert \etal~\cite{Ilbert2024}}
\vskip 0.15in
\begin{small}
\begin{sc}
\begin{tabular}{ccc}
\toprule
\textbf{} & \textbf{LATST} & \textbf{SAMformer} \\
\midrule
MSE Error Percentage & 107.17\% & 100.00\%\\
Average Size of Parameter& 271.4K & 173.2K\\
\bottomrule

\label{tab:overall_imp_param}
\end{tabular}
\end{sc}
\end{small}
\end{center}
\vskip -0.1in
\end{table}

Preliminary results from the ablation study, as shown in \cref{tab:ab_relu}, indicate that the inclusion of the PReLU activation function leads to a significant improvement on the ETTm2 dataset, while providing only marginal enhancements for the other datasets. This observation suggests that the performance gains associated with PReLU may not arise solely from its inherent properties; rather, it appears that the ReLU activation function may hinder the performance of the LSEAttention model due to the numerical instability introduced by its sharp activation characteristics.

To further investigate the contribution of the Log-Sum-Exp trick, we conducted an experiment comparing LATST with a Transformer model that replaces the GeLU activation function.

As illustrated in \cref{tab:ab_gelu}, the LATST model utilizing the Log-Sum-Exp trick significantly outperforms the Transformer model, despite a notable decline in performance compared to the original configuration. This indicates that the Log-Sum-Exp trick plays a crucial role in maintaining the effectiveness of the LATST architecture, even in the absence of the GeLU activation function. The improvement achieved through the Log-Sum-Exp trick is attributed to its ability to stabilize gradient computations during training, thereby facilitating the learning of more complex patterns.

These findings not only elucidate the effectiveness of various components within the LATST architecture, but also provide valuable insights for future enhancements. We aim to explore optimal configurations of these modules to further refine the model's performance and expand its applicability to more complex time series forecasting challenges.

\section{Conclusion and Future Work}
In this paper, we present an innovative solution to mitigate the entropy collapse phenomenon observed in Transformers applied to time series forecasting. Our empirical results confirm the effectiveness and robustness of the proposed LATST approach inty of forecasting contexts. Although we do not exhaustively explore all potential causes of entropy collapse, we introduce a combination of novel modules specifically combined to address this issue.

Looking ahead, future research will focus on a comprehensive investigation of the remaining factors contributing to entropy collapse. This exploration aims to identify additional modifications and enhancements that could further improve the stability and performance of Transformer models in time series forecasting.

In summary, we believe that our work not only addresses critical challenges associated with the entropy collapse phenomenon but also sets the stage for future advancements in time series forecasting methodologies.


% \bibliography{example_paper}
% \bibliographystyle{icml2025}

%%%%%%%%%%%%%%%%%%%%%%%%%%%%%%%%%%%%%%%%%%%%%%%%%%%%%%%%%%%%%%%%%%%%%%%%%%%%%%%
%%%%%%%%%%%%%%%%%%%%%%%%%%%%%%%%%%%%%%%%%%%%%%%%%%%%%%%%%%%%%%%%%%%%%%%%%%%%%%%
% APPENDIX
%%%%%%%%%%%%%%%%%%%%%%%%%%%%%%%%%%%%%%%%%%%%%%%%%%%%%%%%%%%%%%%%%%%%%%%%%%%%%%%
%%%%%%%%%%%%%%%%%%%%%%%%%%%%%%%%%%%%%%%%%%%%%%%%%%%%%%%%%%%%%%%%%%%%%%%%%%%%%%%
% \newpage
% \appendix
% \onecolumn

% You can have as much text here as you want. The main body must be at most $8$ pages long.
% For the final version, one more page can be added.
% If you want, you can use an appendix like this one.  

% The $\mathtt{\backslash onecolumn}$ command above can be kept in place if you prefer a one-column appendix, or can be removed if you prefer a two-column appendix.  Apart from this possible change, the style (font size, spacing, margins, page numbering, etc.) should be kept the same as the main body.
%%%%%%%%%%%%%%%%%%%%%%%%%%%%%%%%%%%%%%%%%%%%%%%%%%%%%%%%%%%%%%%%%%%%%%%%%%%%%%%
%%%%%%%%%%%%%%%%%%%%%%%%%%%%%%%%%%%%%%%%%%%%%%%%%%%%%%%%%%%%%%%%%%%%%%%%%%%%%%%

\end{document}